\documentclass[a4paper, twoside, 12pt]{report}
\usepackage[
   left=25mm,
   right=23mm,
  top=20mm,
  bottom=20mm,
  headheight=15pt
]{geometry}

\usepackage{fancyhdr}
\usepackage{fancyvrb}
\usepackage{titlesec}
\usepackage{framed, color}
\usepackage{graphicx}
\usepackage{url}
\usepackage[numbers, sort]{natbib}
\usepackage{amsfonts}
\usepackage{amssymb}
\usepackage{amsmath}
\usepackage{comment}
\usepackage{mathtools}
\usepackage{sectsty}
\usepackage{afterpage}
\usepackage[]{hyperref}
\usepackage{longtable}
\usepackage[shortlabels]{enumitem}

\newenvironment{myenumerate}
{ \begin{enumerate}[nosep] }
{ \end{enumerate}                  }

\titleformat{\chapter}[display]
  {\normalfont \centering \bfseries}{}{0pt}{\Huge}

\graphicspath{{images/}}

\newenvironment{dedication}
  {\clearpage           %
   \thispagestyle{empty}%
   \vspace*{\stretch{1}}%
   \itshape             %
   \center          %
  }
  {\par %
   \vspace{\stretch{3}} %
   \clearpage           %
}

\newcommand\frontmatter{%
    \cleardoublepage
  \pagenumbering{roman}}

\newcommand\mainmatter{%
  \pagenumbering{arabic}}

\newcommand\backmatter{%
  \if@openright
    \cleardoublepage
  \else
    \clearpage
  \fi
   }

\title{\textbf{Evaluation Function Approximation for Scrabble}}
\author{Bachelor Thesis Report \\ \emph{submitted in partial fulfillment of requirements for the degree of}\\
        \\
        {Bachelor of Technology}\\
        \\
        \emph{by}\\
        \\
		{Rishabh Agarwal}\\
        {Roll No : 140050019}\\
        \\
        \emph{under the guidance of}\\
        \\
		{Prof. Shivaram Kalyanakrishnan}\\
        \\\\
        \includegraphics[height=3.5cm]{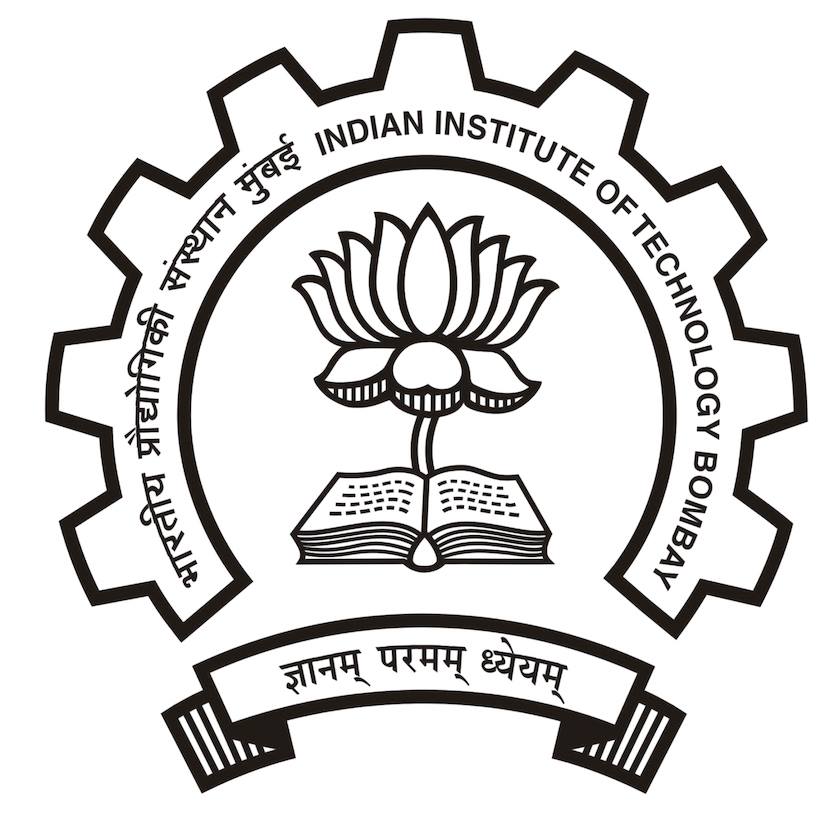}\\
        \\
		{Department of Computer Science and Engineering}\\
        {Indian Institute of Technology Bombay}\\
        {Mumbai 400076, India}\\
}

\begin{document}
\frontmatter
\linespread{1}

\maketitle

\begin{dedication}
To my parents, for their endless support \\
and for teaching me the value of hard work and discipline.
\end{dedication}

\renewcommand{\abstractname}{Abstract}
\begin{abstract}
 \addcontentsline{toc}{chapter}{Abstract}
The current state-of-the-art Scrabble agents are not learning-based but depend on truncated Monte Carlo simulations and the quality of such agents is contingent upon the time available for running the simulations. This thesis takes steps towards building a learning-based Scrabble agent using self-play. Specifically, we try to find a better function approximation for the static evaluation function used in Scrabble which determines the move goodness at a given board configuration. In this work, we experimented with evolutionary algorithms and Bayesian Optimization to learn the weights for an approximate feature-based evaluation function. However, these optimization methods were not quite effective, which lead us to explore the given problem from an Imitation Learning point of view. Currently, we are trying to imitate the ranking of moves produced by the Quackle simulation agent using supervised learning with a neural network function approximator which takes the raw representation of the Scrabble board as the input instead of using only a fixed number of handcrafted features.

\end{abstract}

\clearpage
\thispagestyle{empty}
\topskip0pt
\vspace*{\fill}
\vspace{-5cm}
\begin{center}
\addcontentsline{toc}{chapter}{Acknowledgement}
\bf{Acknowledgement}
\end{center}
I must acknowledge a great debt to my advisor, Professor Shivaram Kalyanakrishnan, for his constant advice has been the guiding force throughout and without which, this work would never have been possible. I also thank him for all his forbearance and wise professional counsel multiple times in the last few months. 
\\ \\
Rishabh Agarwal
\vspace*{\fill}

\tableofcontents
\raggedbottom

\mainmatter

\setlength{\parindent}{2em}
\setlength{\parskip}{0.5em}

\pagebreak

\chapter{Introduction}

\section{Scrabble}

Scrabble\cite{wiki:scrabble} is a board game, where two players alternate on forming words on a 15x15 board by placing tiles bearing a single letter onto a board. The tiles must form words which, in crossword fashion, read left to right in rows or downwards in columns, and be defined in a standard dictionary or lexicon. Also, at least one of the tiles must be placed next to an existing tile on the board. The board is also marked with ``premium" squares, which multiply the number of points awarded for a given word.

The game begins with a total of 100 tiles, placed in a bag in which the letters are invisible.
The set of tiles that a player holds is called the rack and each player starts off with drawing seven tiles each from the bag. The set of tiles left after a player has moved is called the rack leave and for every letter placed on the board a replacement is drawn from the bag. The game ends when one player is out of tiles and no tiles are left to draw, or both players pass. The interested reader can find more details about the rules of Scrabble \href{http://www.scrabblepages.com/scrabble/rules/}{here}.

\begin{figure}[ht]
\centering
\includegraphics[scale=0.4]{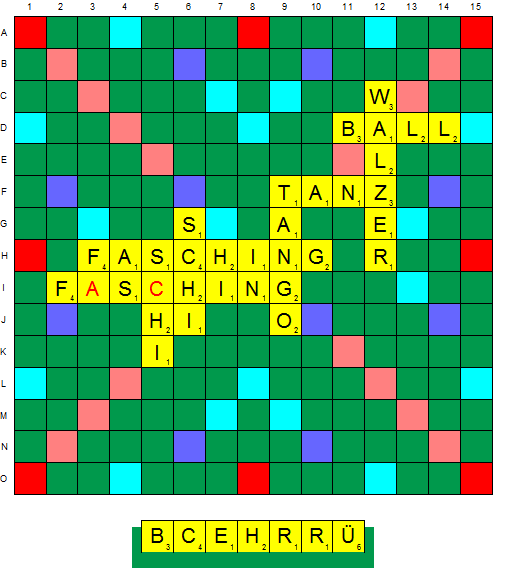}
\caption{An example Scrabble board with the rack of a particular player. Note that the squares other than the premium squares are colored green while other colors correspond to different type of premium squares. Source: \url{http://www.bilder-hochladen.net/files/big/jcry-7r-bf1b.png}}
\label{fig:scrabble_board}
\end{figure}

\section{Why Scrabble?}

Scrabble is a game of imperfect information, i.e. the current player is unaware about the rack of the opponent player, making it very hard to guess the opponent's next move until the end of the game. Also, there is inherent randomness present in Scrabble as random letters are drawn from the bag to the current player's rack at each round. The state space in Scrabble is also quite complex due to the tiles being marked with specific letters as opposed to being black and white. All these factors contribute to the difficulty of creating a perfect AI agent for Scrabble.

\section{Outline}
As a prerequisite to begin our exposition, we provide an overview of current state-of-the-art Scrabble agents with an high level description of the techniques employed in these agents in Chapter \ref{chap2}. Chapter \ref{chap3} provides a detailed discussion of our approach towards the improving the evaluation function using black-box optimization methods. In Chapter \ref{chap4}, we shed light on our current framework involving supervised learning for approaching the problem. We conclude with a brief page of summary and directions for future work in Chapter \ref{chap5}.

\chapter{Background} \label{chap2}

In this chapter, we cover the necessary background required for appreciation of this work. We begin by describing the main components of the Scrabble agent Maven followed by a high level description of Quackle's artificial intelligence. Quackle is the current state-of-the-art Scrabble agent and is based on Maven. Our work builds upon the open source Quackle agent and utilizes its \href{https://github.com/quackle/quackle}{code} available on the Internet.

\section{Maven} \label{maven}
Maven's\cite{sheppard2002world} game play is sub-divided into 3 phases: 
\begin{enumerate}
\item \textbf{Mid-game}: This phase lasts from the beginning of the game up until there are 9 or fewer tiles left in the bag. In this phase, all possible moves from the given rack are generated followed by sorting using some simple heuristics. The most promising moves among these moves are evaluated using 2-ply Monte-Carlo simulation where both players are assumed to draw tiles from the bag in each turn, and are assumed to play the best move based on the aforementioned heuristics. The move which is evaluated to be best after the simulations is played during the game.
\item \textbf{Pre-endgame}: This phase is almost similar to the mid-game phase and is designed to attempt to yield a good end-game situation.
\item \textbf{Endgame}:  During this phase, there are no tiles left in the bag and thus, Scrabble becomes a game of perfect information. Maven uses the $B^\star$ search algorithm to analyze the game tree during the endgame phase.
\end{enumerate}

\section{Quackle}
\begin{figure}[ht!]
  \includegraphics[scale=0.5]{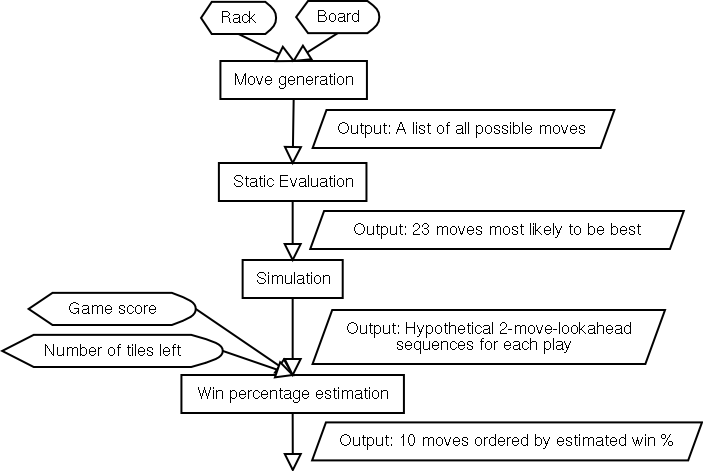}
  \centering
  \caption{Quackle's AI in a Block Diagram. Source: \url{http://people.csail.mit.edu/jasonkb/quackle/doc/blockdiagram.png}}
  \label{fig:quackle_block}
\end{figure}

Quackle\cite{katz-brown_o'laughlin} uses a similar approach as used by Maven. Quackle's heuristic function\footnote{Henceforth, shall be referred to as the static evaluation function} used in the mid-game phase is a sum of the move score and the value of the rack leave. This leave value is computed using a precomputed database: it favors both the chance of playing a Bingo on the next turn as well as leaves that garnered high scores when they appeared on a player's rack in a large sample of Quackle versus Quackle games. The win-percentage of each move is estimated and the move with the best win-percentage is played. Figure \ref{fig:quackle_block} presents a high level flow chart of the algorithm used in Quackle.

These are mainly two types of players present in the Quackle open source \href{https://github.com/quackle/quackle}{code} which was used for our experiments:
\begin{enumerate}
\item \textbf{Speedy Player}: The player only uses only static evaluations throughout the game and no Monte-Carlo simulations. The move predicted to be the best by the static evaluation function is played at a given board configuration.

\item \textbf{Championship Player}: This player uses simulations in addition to static evaluations along with a perfect endgame player. This player can be provided with a particular amount of think-time\footnote{The win-rate of second player when the first player is a ``Twenty Second Championship Player'' i.e. it has a think-time of 20 seconds/move and the second player is a ``Speedy Player'' is $\approx \textbf{41.3\%}$ calculated using 50000 games.} to run the truncated Monte-Carlo simulations for a single turn. For example, ``Five Minute Championship Player" can take as long as five minutes to decide upon the move to play in a single turn.
\end{enumerate}

Please note that the win-rate of second player when both players are Speedy Players is $\approx \textbf{43.8\%}$, calculated using 50000 self-play games.

\chapter[Evaluation function]{Improving the evaluation function} \label{chap3}
The static evaluation function used in Quackle takes a list of moves as input and outputs a ranked list of moves by a rough guess of how strong they are. After applying the static evaluation function to get a small number of ranked moves, the Quackle agent further uses truncated Monte-Carlo simulations to decide the best move among this moves. Our work deals with improving the static evaluation function in Quackle so that the performance of the ``Speedy Player'' can be improved significantly. We hypothesize that this improved evaluation function would also potentially decrease the simulation time for the ``Championship Player" by predicting a smaller number of moves likely to be best at a given board configuration in Scrabble.

We extensively experimented with evolutionary approaches and Bayesian Optimization\cite{brochu2010tutorial} to improve the evaluation function using self-play. Both of these approaches requires a fitness function\cite{wiki:fitness_func} to estimate the goodness of a given evaluation function. We will first discuss some of the technical details regarding the fitness function followed by a discussion of the above-mentioned approaches.

\section{Fitness Function} \label{fitness}

\subsection{True Fitness Function}
The true fitness function, $f_{true}$ for a given evaluation function is given by the win-percentage of the Scrabble agent using the given evaluation function against another agent (usually Speedy Player) using another fixed evaluation function. The win-percentage should be estimated using a large number of games, in order to keep the error-bars around the win-percentage small, leading to a reasonable estimate of the function $f_{true}$.

\phantomsection \label{par:error_bars}
Using Hoeffding's inequality\cite{wiki:hoeffding}, the true mean of a Bernoulli random variable that has been sampled N times will lie, with probability at least 1 - $\delta$, within the empirical mean $\pm \sqrt{\log(2/\delta)/(2*N)}$. The win-rate approximated using N self-play games is estimated correctly with an error bound given by the above inequality. Using $ 1 - \delta = 99.9\% $, and N = 5000 leads to a error bar of $\pm  0.0275$. Further, increasing N to 50000 games still leads to an error of $\approx 0.86\%$. We can achieve slightly tighter error bounds based on KL Divergence\cite{kaufmann2013information} given by equations \ref{eq1} and \ref{eq2}.
\begin{align}
upper\_error\_bound(\delta, N) &= max\ \{q \in [p\star, 1] : N * KL(p\star, q) \leq  \log(1/ \delta)\} \label{eq1}\\
lower\_error\_bound(\delta, N) &= min\ \{ q \in [0, p\star] : N * KL(p\star, q) \leq \log(1/ \delta) \} \label{eq2}
\end{align}
where $p\star$ denotes the true mean of the Bernoulli and N is the number of self-play games. These equations also a error of $\approx 0.8\%$ for $N = 50000$ games for  $1 - \delta = 99.9\%$. 

\subsection{Other fitness functions}
\label{par:f_sim}
The idea of imitating the simulation agent (``Championship Player'') to improve the static evaluation function also seemed quite reasonable and in order to explore this idea more, we experimented with the fitness function $f_{sim}$, which utilized the ranking of the moves by the ``Five-Minute Championship Player".  We generated $\approx$ 70000 board configurations, and for each board configuration \textit{b}, we calculated a ranked list $L_b$ of size min$\{10, M_b\}$ moves out of all possible moves $M_b$ using the ``Championship Player". For each board configuration, a list of top min$\{ 10, M_b\}$ was generated using the given static evaluation function. The player was provided a score of $1/k$ where k is the index of the best move in $L_b$ among those 10 moves. The function $f_{sim}$ is given by the sum of all such scores.

We also tested the fitness function $f_{score}$ given by the sum of final score differences for a ``Speedy Player'' over all games played between a fixed Quackle player and the ``Speedy Player'' which uses a static evaluation function. However, the preliminary results\footnote{$f_{true}$ obtained a win-rate of 46\% as compared to the win-rate of 44.5\% obtained by $f_{score}$ using CMA-ES with 6 generations and a population size of 50. Please note that these results are correct to within $\pm 2.5\%$ as they were obtained using fitness function evaluation over 5000 games only .} which we obtained by using $f_{score}$ were worse as compared to using $f_{true}$ and therefore, we didn't use $f_{score}$ in any further experiments.

\section{CMA-ES}
\begin{figure}[ht]
\centering
\includegraphics[scale=0.7]{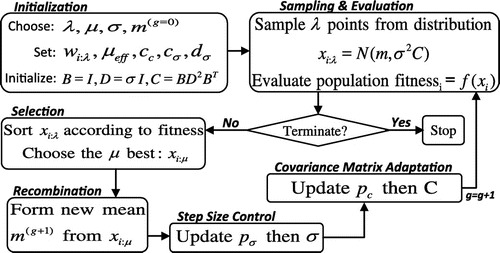}
\caption{CMA-ES flowchart. In each iteration(generation) step, a weighted combination of the $\mu$ best out of $\lambda$ new candidate solutions is used to update the distribution parameters $p_\sigma$, $p_\mu$, $\vec{C}$.
Here, $\mathbb{N}$ represents a normal distribution and $f$ is the fitness function. Source: \url{http://ascelibrary.org/cms/attachment/9942/197821/figure3.gif}}
\label{fig:cma_es}
\end{figure}

We used an evolutionary algorithm\cite{spears1993overview}, called CMA-ES (Covariance Matrix Adaptation Evolution Strategy)\cite{hansen2016cma}, to find the optimal weights for a feature-based evaluation function. An evolutionary algorithm is broadly based on the principle of biological evolution: In each generation, $\lambda$ new individuals are generated by stochastic variation of the current parental individuals. Then, the top $\mu$ individuals out of $\lambda$ are selected to become the parents in the next generation based on their fitness function value $f$. In this manner, over the generation sequence, individuals with better and better $f$-values are generated. Figure \ref{fig:cma_es} presents a high level overview of the CMA-ES algorithm.

For our experiments, we ran 4-6 generations of CMA-ES
with $\lambda = 25$ and $\mu = 13$ agents using the fitness function $f_{true}$ calculated through $N = 50000$ self-play games between two ``Speedy Players'' as described in section \ref{fitness}.

\subsection{Linear Evaluation Function} \label{linear_eval}
The linear static evaluation function scores a move based on some state-action features $\phi$ generated using the current board configuration and the move to be evaluated. The computed score is a dot product of the feature values with a weight vector. Note that the state-only features are not used as they will be same for all the moves at a given state and thus would not affect the ranking using a linear function. The move with the highest value of the evaluation function is played. The feature set $\phi$ consists of the following features:
\begin{enumerate}
\setlength\itemsep{0.03em}
\item Move Score: The score we obtain by playing the given move on the board.
\item Leave Value: A precomputed value for each rack leave generated by the Quackle code
\item Leave Playability: A value calculated for each rack leave which calculates the expected value of the move score that can be obtained after sampling letters from the bag to complete the rack and using that rack to form a word
\item Difference in consonants and vowels left on the rack
\item Number of blanks left on the rack after playing the move
\end{enumerate}
The current static evaluation function in Quackle, $e_{quackle}$ only uses the top two features mentioned above i.e. the move score and leave value, with a weight of $1.0$ for each of the two features.

\begin{figure}[!ht]
\centering
\includegraphics[scale=0.8]{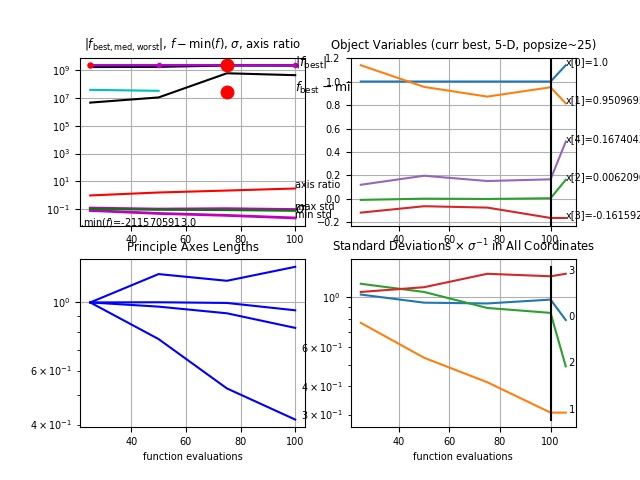}
\caption{Result of applying CMA-ES using only the first 5 features for a linear evaluation function with the weight of first feature always fixed to 1. The upper left shows the variation in the function $f$ $\propto$ $-f_{true}$ with $f_{best}$ being the least $f$-value in a given generation. The lower figures show the square root of eigenvalues (left) and of diagonal elements (right) of the covariance matrix $C$. The actual sample distribution has the covariance matrix $\sigma^{2}C$.}
\label{fig:cma_es_5}
\end{figure}

Figure \ref{fig:cma_es_5} shows the results we obtained after running CMA-ES with above-mentioned features. As shown in the upper left plot in this figure , the value of the $f_{best}$ - min($f$) is increasing as the generation number increases except for the last generation. The final weights with the min($f$) (or max($f_{true}$)) results in a win-percentage of $\approx\textbf{44.1\%}$\footnote{This is an improvement of 0.3\% over $e_{quackle}$} calculated over 50000 games for the second player.

We also experimented with other features such as the length of the rack leave and features corresponding to the board configuration changes after playing a move. For example, when a player plays a move on the board, new openings (positions which can be used to play a valid move) are created on the board for the opponent player. Some of these openings can also utilize premium squares. We experimented with the feature corresponding to number of such openings for each kind of premium square, leading to a total of $4$ such features. However, these new features deteriorated the maximum fitness value obtained using CMA-ES and therefore, were skipped for further experiments. The fact that the above mentioned features didn't lead to any improvement is surprising, and we believe that this was due to the limited representation ability of the linear evaluation function.

Using CMA-ES, $f_{sim}$, as described in section \ref{par:f_sim} also didn't lead to improvement with the above mentioned features\footnote{Surprisingly, this experiment resulted in a win-rate of 41.8\% when $f_{sim}$ was evaluated using 50000 games.}. The baseline $f_{sim}$-value for $e_{quackle}$ indicates that the best move generated by the ``Championship Player" is expected to lie withing the top 5 moves predicted by $e_{quackle}$.\\ \label{par:e_quackle}

\subsection{Non-Linear Evaluation function} \label{nn:eval_fn}
Since a linear model has limited representation ability, we also experimented with evaluation functions represented by a neural network with 1-2 hidden layers and non-linear activation functions such as tanh and ReLU\cite{nair2010rectified}. The neural network is only used to introduce the non-linearity in a structured manner. State-only features pertaining only to the board such as differences of vowels and consonants on the board, number of blanks on board etc. were also used as input to the network, in addition to state-action features mentioned in section \ref{linear_eval}. The results from this experiment were not encouraging and the given class of non-linear functions were not able improve upon $e_{quackle}$.

\section{Bayesian Optimization}
Bayesian optimization\cite{brochu2010tutorial} is a method of optimization of expensive cost functions without calculating derivatives. This kind of optimization employs the Bayesian technique of setting a prior over the objective function and combining it with evidence to get a posterior  function. This allows an utility-based selection of the next observation using an acquisition function\cite{snoek2012practical} that determines what the next query point should be. The acquisition function takes into account both sampling from areas of high uncertainty as well as areas likely to offer improvement over the current best observation.

Since our fitness functions are noisy (as only finite number of games are used to compute them), the idea of applying Bayesian optimization to find the optimal static evaluation function looked promising. However, this experiment also didn't result in a significant improvement over the current win-rate of $43.8\%$ obtained by $e_{quackle}$.

\section{Multiple evaluation functions}
Instead of using the same evaluation function for all stages\footnote{Refer to section \ref{maven} for detailed information about the different phases.} of the game, it seemed plausible to us that a combination of evaluation functions could work better for different game stages in Scrabble. This idea has been previously employed successfully in the game of Robot Soccer\cite{macalpine2012ut} and already been proposed in the context of Scrabble\cite{romero2009} as well. Keeping the evaluation function weights fixed for the early-game\footnote{Early-game was defined as the phase of the game when the number of tiles in the bag were $\geq$ 80.} and endgame phases, we only evolved the weights pertaining to the mid-game where the mid-game was defined to be the phase of game when the number of tiles in the bag were between 20 to 80. This experiment also proved to be futile and didn't result in any improvement over the function $e_{quackle}$.

\chapter[Imitation Learning]{Imitation Learning in Scrabble} \label{chap4}

Imitation learning\cite{hussein2017imitation} (or Learning from Demonstrations\cite{argall2009survey}) aims to mimic expert behavior in a given task. In this technique, an agent is trained to perform a task using demonstrations by learning a mapping between observations and actions. This approach has been quite successful in AlphaGo\cite{silver2016mastering} where deep neural networks were trained by a combination of supervised learning from human expert games (imitation Learning) and reinforcement learning from games of self-play.

The failure of the black-box optimizations methods made us rethink our approach for improving the static evaluation function. The observation made at the end of section \ref{par:e_quackle} that the top move predicted by current static evaluation function in Scrabble lies within the top 5 moves of the best simulation Quackle agent motivated us to actually learn the evaluation function instead of evolving it using a noisy fitness function. Our main idea is to imitate the ranking produced by the Quackle simulation agent using only the static evaluation function. 

\section[Learning to Rank]{Learning to Rank} \label{par:lr_rank}
We initially experimented with a neural network approximator $G$ as the static evaluation function with input as a move represented by the various state-action features (including state-only features as well) described in section \ref{nn:eval_fn}. We trained this neural network using a fixed dataset $D$ of state-action pairs with $\approx$ 70000 board configurations. For each board configuration $b$, the Quackle ``Five Minute Championship Player'' provides a sorted list $L_b$ of size min\{10, $M_b$\}, where $M_b$ is the list of all possible moves at configuration $b$. The dataset D is further split into training and validation datasets $D_{train}$ and $D_{val}$ with a ratio of 97:3.

\begin{figure}[ht]
\centering
\includegraphics[scale=0.6]{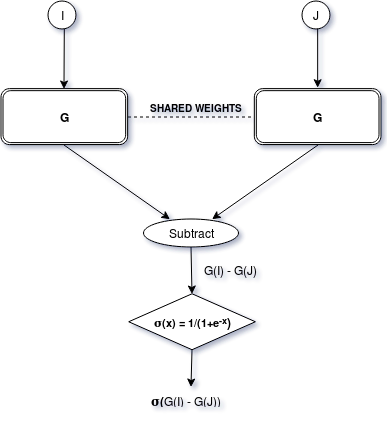}
\caption{Architecture of the network $G_{pair}$. Given two moves $I$, $J$ for a given board configuration(state), the network $G_{pair}$ takes the tuple ($I$, $J$) as input and outputs $\sigma(G(I) - G(J))$. $G_{pair}$ is trained using a binary cross-entropy loss where the label for pair $I$, $J$ is 1 if the move $I$ was ranked above move $J$ by the Quackle agent, otherwise the label is 0.}
\label{fig:nn_arch}
\end{figure}

Figure \ref{fig:nn_arch} presents the Rank-net\cite{burges2005learning} architecture for $G_{pair}$ which invokes the network $G$. For each board configuration $b$ in dataset $D$, the network $G_{pair}$ is provided with pair of moves ($L_{b}[1]$, $L_{b}[k]$) where $k \in \{2, 3, .. ,|L_{b}|\}$ where L[i] denotes the $i^{th}$ element of the list $L$. The output label, $y_{output} =  1$ for all such pairs. The binary cross-entropy loss function used for training is given by $$\sum_{n=1}^{|D_{train}|} \sum_{i = 1}^{|L_{D_n}|} -\log(\sigma(G(L_{D_{n}}[1]) - G(L_{D_{n}}[i])))$$

However, results from this experiment were not very encouraging. The function approximator $G$ was only able to achieve a slightly better win-rate than $e_{quackle}$. Further, the validation accuracy achieved using the ranking loss was close to the baseline accuracy($\approx 73\%$) produced by $e_{quackle}$. We believe that the limited representation ability of the features used for this experiment was the limiting factor in this experiment. 

\section{Raw Board Representation}
The problem of designing good features for the game of Scrabble is quite important. The goodness of a board configuration depends both on spatial arrangement and on the presence or absence of sets of letters and ultimately needs consultation with the Scrabble dictionary as well. However, we think that combining these dimensions into a small number of discriminative signals is not an easy task. In order to tackle this problem, we decided to use the raw board representation of the Scrabble board as input to our evaluation function instead of using only a fixed number of hand-coded features.

AlphaGo Zero\cite{silver2017mastering} used the raw board representation as input to neural networks trained using self-play and mastered the game of Go without using any human knowledge. Unlike Go, which only has white and black stones, Scrabble has 27 letters, each with a different value, and with combinations of letters having different a value from the sum of the constituents. This makes the Scrabble board a much harder proposition for a convolutional neural net(CNN\cite{lecun2015deep})\footnote{Henceforth, convolutional neural net is abbreviated as CNN} to model well. However, given the enormous success of AlphaGo Zero, we decided to further investigate the idea of using raw board representations for Scrabble.

A $15\times15$ Scrabble board $b$ is encoded into a $15\times15\times28$ feature vector $v(b)$ where the third dimension pertain to the different type of features used in our encoding of the Scrabble board. For a particular feature, each board position is a given a value. We used the following features for our encoding:
\begin{enumerate}
\setlength\itemsep{0.01em}
\item Whether a position is blank or not
\item Whether a position contains a particular alphabet, leading to a total of 26 features for the letters A-Z
\item The score of the tile placed on a position. All positions not containing any tiles are given a score of zero except the premium scores which are given a score of -1.
\end{enumerate}

The move at a given board configuration $b$ is provided as input using the concatenation of the feature vectors $v(b)$ and $v(b')$ (let's call it \textit{input1}) where $b'$ is the configuration obtained after playing the given move on $b$. In addition to \textit{input1}, the two features used in $e_{quackle}$ (denoted by \textit{input2}) also provided as input to our evaluation function. The neural network $G$ consists of many residual blocks\cite{he2016deep} of convolutional layers with batch normalization\cite{ioffe2015batch} and ReLU activations and is trained in a similar manner as described by figure \ref{fig:nn_arch}.

Specifically, $G$ consists of a single convolutional block followed by 2 residual blocks and a final block. 
The convolutional block applies the following modules sequentially to \textit{input1}:
\begin{myenumerate}
\item A convolution of 16 filters of kernel size $3\times3$ with stride 1
\item Batch normalization
\item A rectifier nonlinearity
\end{myenumerate}
Each residual block applies the following modules sequentially to its input:
\begin{myenumerate}
\item A convolution of 16 filters of kernel size $3\times3$ with stride 1
\item Batch normalization
\item A rectifier nonlinearity
\item A convolution of 16 filters of kernel size $3\times3$ with stride 1
\item Batch normalization
\item A skip connection that adds the input to the block
\item A rectifier nonlinearity
\end{myenumerate}
The final block applies the following modules sequentially to its input:
\begin{myenumerate}
\item A convolution of 1 filter of kernel size $1\times1$ with stride 1
\item Batch normalization
\item A rectifier nonlinearity
\item A fully connected linear layer to a hidden layer of size 32
\item A rectifier nonlinearity
\item A fully connected linear layer to a hidden layer of size 4
\item A rectifier nonlinearity
\item A concatenation layer which combines the previous layer input and \textit{input2}
\item A fully connected linear layer to a scalar
\end{myenumerate}

The number of weight parameters in $G$ were approximately 1/20 times the number of training example pairs used to train $G$. 

\begin{figure}[h!]
\centering
\includegraphics[scale=0.5]{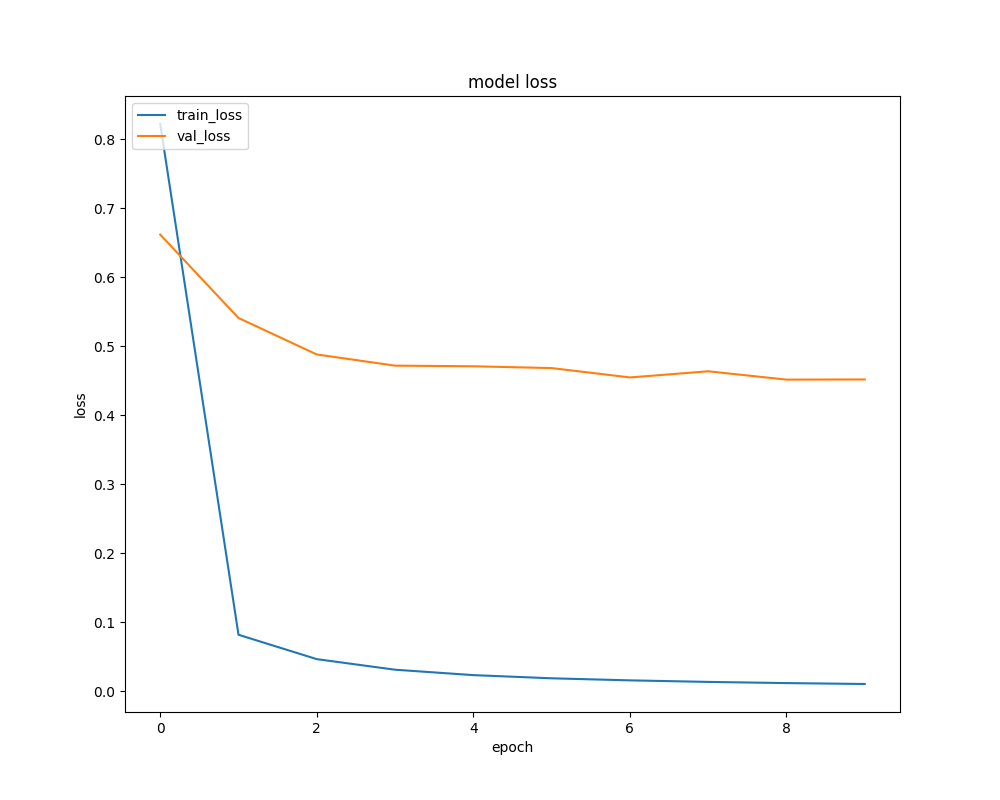}
\caption{Training and validation loss against the number of training epochs. It can be observed that the training loss continually decreases with increasing number of epochs. The validation loss follows the same trend as the training loss, however, it stabilizes to a much higher loss value indicating overfitting.}
\label{fig:nn_raw}
\end{figure}

Figure \ref{fig:nn_raw} shows the learning curves for the network $G$. This approach led to an validation accuracy of $\approx 85.4\%$ on $D_{val}$ which was not used for training. Given these initial results, this approach looks very promising and future work may involve exploring it further.

\chapter{Conclusion and Future Work} \label{chap5}
In this work, we experimented with various black-box optimization methods including CMA-ES and Bayesian Optimization to find a better feature-based static evaluation function than the one currently employed in Quackle. However, the failure of methods mentioned above for both linear and non-linear evaluation functions suggest that the limited representation ability of these feature-based techniques was a possible cause of their failure. This hypothesis is also consolidated by the poor performance of the feature-based Imitation Learning approach described in Section \ref{par:lr_rank}. 

The Imitation Learning methodology utilizing raw board representations shows some signs of success and should be explored further. These are some of the possible improvements over the current approach which can be further investigated:
\begin{enumerate}
\item The loss function currently used corresponds to comparing only the top-ranked move with other moves. However, this loss function completely discards the comparison between other moves. But, to utilize our data more efficiently, these comparisons should also be used for training the neural network. A loss function incorporating all possible comparisons in a weighted manner may lead to better results. This can be achieved by implementing the LambdaRank\cite{burges2007learning} instead of our current RankNet framework.

\item Our current approach involving supervised learning can be complemented by techniques like Dataset Aggregation(DAGGER\cite{ross2011reduction}), where we try to correct our predictions on new board configurations using the ranking produced by the ``Championship player'' while trying to preserve the correctness of our predictions on the old dataset. The reinforcement learning algorithm introduced by the AlphaGo Zero paper can also be implemented for the given problem in hand.
\end{enumerate}

\bibliographystyle{unsrt}
\addcontentsline{toc}{chapter}{Bibliography}
\bibliography{report}

\end{document}